\documentclass{article}

\usepackage[preprint]{neurips_2022}

\usepackage{booktabs}
\usepackage{multirow}

\usepackage[utf8]{inputenc} 
\usepackage[T1]{fontenc}    
\usepackage[hidelinks]{hyperref}       
\usepackage{url}            
\usepackage{booktabs}       
\usepackage{amsfonts}       
\usepackage{nicefrac}       
\usepackage{microtype}      
\usepackage[dvipsnames]{xcolor}         
\usepackage{natbib}
\usepackage{graphicx}

\title{Large language models can segment narrative events similarly to humans}

\author{%
  Sebastian ~Michelmann \\
  Princeton Neuroscience Institute\\
  Princeton University\\
  Princeton, NJ 08540 \\
  \texttt{s.michelmann@princeton.edu} \\
  \And
  Manoj ~Kumar \\
  Princeton Neuroscience Institute\\
  Princeton University\\
  Princeton, NJ 08540 \\
  \texttt{} \\
  \And 
  Kenneth A. ~Norman \\
  Princeton Neuroscience Institute\\
  Princeton University\\
  Princeton, NJ 08540 \\
  \texttt{} \\
  \And 
  Mariya Toneva \\
  Max Planck Institute for Software Systems\\
  66123, Saarbrücken, Germany \\
  \texttt{mtoneva@mpi-sws.org} \\
}

\begin{document}

\newcommand{\TODO}[1]{\textcolor{orange}{TODO: #1}}
\newcommand{\MT}[1]{\textcolor{cyan}{MT: #1}}
\newcommand{\SM}[1]{\textcolor{purple}{SM: #1}}
\newcommand{\KN}[1]{\textcolor{red}{KN: #1}}

\newcommand\blfootnote[1]{%
  \begingroup
  \renewcommand\thefootnote{}\footnote{#1}%
  \endgroup
}

\maketitle

\blfootnote{Code available at \href{https://github.com/s-michelmann/GPT\_event\_segmentation}{\texttt{\textcolor{Turquoise}{https://github.com/s-michelmann/GPT\_event\_segmentation}}}.}

\begin{abstract}
  Humans perceive discrete events such as "restaurant visits" and "train rides" in their continuous experience. One important prerequisite for studying human event perception is the ability of researchers to quantify when one event ends and another begins. Typically, this information is derived by aggregating behavioral annotations from several observers. Here we present an alternative computational approach where event boundaries are derived using a large language model, GPT-3, instead of using human annotations. We demonstrate that GPT-3 can segment continuous narrative text into events. GPT-3-annotated events are significantly correlated with human event annotations. Furthermore, these GPT-derived annotations achieve a good approximation of the ``consensus'' solution (obtained by averaging across human annotations); the boundaries identified by GPT-3 are closer to the consensus, on average, than boundaries identified by individual human annotators. This finding suggests that GPT-3 provides a feasible solution for automated event annotations, and it demonstrates a further parallel between human cognition and prediction in large language models. In the future, GPT-3 may thereby help to elucidate the principles underlying human event perception. 
\end{abstract}

\section{Introduction}
\label{sec:intro}
Humans perceive events in continuous experience \citep[e.g., "restaurant visits" and "train rides";][]{zacks_event_2007}. This inferred event structure has been shown to play a key role in numerous cognitive functions (see Section \ref{sec:background}). Researchers have studied event cognition extensively in controlled settings where event structure is predetermined by the experimenter \citep[e.g.,][]{dubrow_temporal_2016}, but recently there has been renewed interest in studying event cognition in more ecological settings \citep[see][]{sonkusare_naturalistic_2019}. Studying event cognition with naturalistic stimuli like movies and stories typically involves laborious hand annotation of event boundaries that are often crowd-sourced from large behavioral samples in online experiments  \citep[e.g.,][]{michelmann_moment-by-moment_2021, michelmann_evidence_2022}. Because annotations may vary between participants, those large samples approximate a shared perception by averaging across many individual annotations. Notably, this approach is not scalable as the field becomes interested in studying how people process long naturalistic stimuli
; getting suitably large numbers of participants to read and annotate boundaries in a 15 hour audiobook would be a forbidding endeavor. 

In this work, we propose to leverage recent developments in language models to enable progress in studying event cognition at naturalistic scales. Large language models that have been pretrained to predict the next word over billions of text documents have swept up the natural language processing (NLP) field and led to state-of-the-art performance across NLP tasks. In parallel, cognitive neuroscientists have also begun evaluating these large language models with respect to their alignment with human behavior \citep{caucheteux2021gpt,golan2022testing,kumar_bayesian_2022} and human brain activity during language comprehension \citep{jain2018incorporating, toneva2019interpreting, schrimpf2021neural, caucheteux2020language, goldstein2022shared}, finding many significant correspondences. However, it is still an open question to what extent the inner workings of these large language models reflect human cognition.

The current approach examines whether a powerful large language model, GPT-3 \citep{brown2020language}, can segment events in a similar way to humans. Specifically, we assess its ability to perform event segmentation tasks using only zero-shot prompting (i.e., without any additional fine-tuning or in-context learning). We prompt GPT-3 with a similar (albeit simpler) prompt to what human participants are shown when asked to segment a natural stimulus into meaningful events. Across a variety of stories, we find that GPT-3 is able to perform this task significantly above chance. Moreover, we demonstrate that GPT-derived event boundaries provide a good approximation of the consensus solution (obtained by averaging responses across people); the boundaries identified by GPT-3 are closer to the consensus, on average, than boundaries identified by individual human annotators. This suggests that GPT-3 annotates event boundaries according to a shared human perception that is captured by averaging across large samples of human annotators. Finally, we derive a continuous measure of event probability from the log-probability of the newline character (that was used to indicate a new event) in the model output. This continuous measure of event boundary perception in the model is significantly correlated with the continuous measure of agreement between human raters. This finding provides further evidence that GPT-3 can represent the event structure of a naturalistic story in its predictions, rendering it a feasible tool for obtaining automated event annotations. We make our code publicly available to enable others to easily integrate the proposed approach into their analyses.






\section{Background}
\label{sec:background}
Numerous studies have demonstrated the role of event structure in cognitive phenomena such as working memory \citep{jafarpour_event_2022}, time-perception \citep{bangert_crossing_2020}, and episodic memory \citep{dubrow_influence_2013, michelmann_evidence_2022, sargent_event_2013}. Furthermore, neural correlates of events have been identified in human functional magnetic resonance imaging \citep[fMRI, ][]{baldassano_discovering_2017}, electroencephalography \citep[EEG, ][]{silva_rapid_2019} and intracranial EEG \citep[iEEG, ][]{michelmann_moment-by-moment_2021}. Moreover, a recent study identified "event cells" in rodents: single neurons that represented laps in a maze were invariant to lap-duration \citep{sun_hippocampal_2020}. \citet{brunec_boundaries_2018} and \cite{clewett_transcending_2019} summarize in recent reviews how event boundaries shape behavior and human brain activity. Crucially, a substantial number of event-perception studies have focused on the comprehension of narrative text. In these studies, participants typically read a text \citep{zacks_segmentation_2009, ezzyat_what_2011, pettijohn_narrative_2016}, or listen to a spoken narrative \citep{ zacks_segmentation_2009, michelmann_moment-by-moment_2021} with the instruction to segment the narrative into events. Importantly, individual raters agree substantially on the perception of event boundaries, drawing on an underlying shared perception of long-time-scale dynamics in narratives \citep{chen_shared_2017}.

The ubiquitous influence of event structure on human cognition and neural activity suggests that there may be universal computational principles at work that make event segmentation useful for solving cognitive tasks \citep[e.g.,][]{lu_neural_2022}. \citet{zacks_event_2007} and others \citep[e.g.,][]{shin_structuring_2021} have posited that the segmentation of ongoing experience into discrete events is driven by prediction error. According to this perspective, individuals generate predictions about the current event based on learned statistical information. For instance, we expect to be asked for our ticket on a train ride but would be surprised if the waiter at a restaurant asked us for our ticket instead of offering the menu. When these predictions begin to fail, it is inferred that the current event has ended and that a new event has begun, marking an event boundary. This hypothesized link between prediction and event segmentation suggests that large neural network models, which are likewise concerned with the prediction of upcoming information, may show sensitivity to event structure. In support of this view, recent studies have shown that measures of prediction error derived from large network models are significantly correlated with human annotations of event boundaries \citep{kumar_bayesian_2022, roseboom_perception_2022}.
Currently, however, no method has been designed to produce an automated event segmentation of text. While our method is currently agnostic to the underlying mechanisms that drive event segmentation, we note that it can explain substantial and significant variance in behavior (maximal correlation of $r = .372$) and that our segmentation performs within the expected range of individual human annotators (see Section \ref{sec:results}).

\section{Methods}
\label{sec:methods}

We develop a simple approach to enable GPT-3 to segment narrative stories into meaningful events, utilizing recent success in zero-shot prompting. We validate this approach across three stories of different lengths, and compare the resulting GPT-3 event segmentations with annotations from human participants.

\subsection{Materials}
We tested our approach on narrative stories of three different lengths, which were previously used in psychological and neuroscientific experiments: 1) "Pieman", a short story of seven minutes and $30$ seconds that was presented at a live storytelling event at the Moth \citep{ogrady_2008}, 2) "Monkey in the Middle", a 30-minute-long audio podcast that appeared on "This American Life" \citep{chavis_2017}, 3) "Tunnel Under the World", a 25-minute-long science fiction story that was broadcast on the radio show "X Minus One" \citep{pohl_frederik_tunnel_1956}. 

\paragraph{Preprocessing.} Transcriptions of these auditory stories were used for their segmentation with GPT-3 in this study. Transcriptions were in plain text (UTF-8 encoding), no hyphens were present, and quotation marks and occurrences of ellipses were removed. The fast implementation of the pretrained GPT-2 tokenizer was used to tokenize the text (huggingface.co). The transcripts were aligned to the audio recordings via Penn Phonetics Lab Forced Aligner \citep{yuan_pfa_2008}. This was done to later match behavioral button press responses (that had been recorded in milliseconds from the start of the audio) to positions in the text. 

\subsection{GPT-3 event segmentation}
\paragraph{Prompting.} In order to prompt GPT-3 to segment a continuous story into events, a completion was created with the model version "text-davinci-002" (beta.openai.com/docs/models/). The specific prompt for the model was: "\textit{An event is an ongoing coherent situation. The following story needs to be copied and segmented into events. Copy the following story word-for-word and start a new line whenever one event ends and another begins. This is the story:} ". After this instruction, a segment from the story was inserted. Finally, after a new line, an additional prompt segment was added to refresh the instruction: "\textit{This is a word-for-word copy of the same story that is segmented into events:}". In a variation of the instruction, the qualifier "\textit{long}" was added before every occurrence of the word event in the prompt, with the exception of the first definition. This was done to obtain a coarse-grained event segmentation. 

\paragraph{Sliding window for long stories.} The context length of the GPT-3 model that we used was not long enough to encompass each of the three stories. Therefore, we used a sliding window approach to split the stories into segments that fit into the context length of GPT-3. The length of a story segment was determined based on the number of input and output tokens that fit into the context window (see below for some notes on the width that was used for the context window). The output text that the model generated was then split into events at newline characters. When the final event in a story segment stopped before the end of the full story, the next story segment was taken from the beginning of the final event of the previous segment and continued into the story for the next completion request. The segmentation was complete after all text from a given story had been parsed. 

\paragraph{Hyperparameters.} To increase reproducibility, the \textit{temperature} parameter was set to 0. Additionally, only half of the width of the context window (i.e., $4096/2 = 2048$) was used, because the model output occasionally diverged from a word-for-word copy when longer context windows were used. The total context window was split between prompt and output tokens: The number of instruction tokens in the prompt was first deducted from the $2048$ token window width, and the remaining tokens were then divided equally between tokens that were reserved for the story segment used in the prompt and the number of tokens that were requested in the output by setting the  \textit{max\_tokens} parameter. For the story "Tunnel Under the World", this parameter setting proved to be too restrictive and occasionally led to truncated output; to address this, additional padding of 512 tokens was added to the \textit{max\_tokens} parameter for the "Tunnel" story. 

\paragraph{Continuous segmentation probability.} To obtain a continuous output of story segmentation from the model, the log-probability of the top 5 predicted tokens was requested in the completion. The log-probability of the newline character was used to assess to what degree the model predicted an event boundary at any given token in the story. Whenever the newline character was not among the top 5 predicted tokens, the log-probability was not defined. To interpolate the token-probabilities onto the same time axis as the behavioral data, the onset and offset times of words in the transcript were used. Words from the transcript were converted to tokens with the pretrained GPT-2 tokenizer; whenever a word was converted to more than one token, the interval spanned by its onset and offset was evenly split among the resulting tokens which yielded token-onset and -offset times. Next, all words from a given GPT-3 generated response were tokenized and word embeddings from the GPT-2 Model transformer (GPT2LMHeadModel) were used to map model-output tokens onto the list of aligned tokens from the transcript via dynamic-time-warping \citep[using the fastdtw implementation][distance metric: euclidean distance]{salvador_toward_2007}. The warp path was then applied to the token-onset and -offset times, which resulted in an onset and offset for every token that was generated by GPT-3. Finally, a vector of NaN values of the same length as the time-axis of the story was generated; for those tokens where a newline character was among the top-5 predicted tokens, the value of that vector was set to the log-probability of the newline character in the whole interval between the token's onset and offset time. The outcome of this transformation was a vector of the same length as the behavioral response vector from the human participants; the vector held the log-probability of the newline character whenever it was among the top-5 predicted tokens and NaN otherwise (note that NaN was used as a placeholder, these values were later interpolated). 

\subsection{Human event segmentation}
To assess the quality of GPT-3's automated event segmentation, we compare the event boundaries segmented by GPT-3 to those annotated by large numbers of human participants. Data from the human behavioral event segmentation tasks were previously analyzed and published, and were made available by the authors upon our request. In this subsection, we describe the details of these datasets, which are summarized in Table \ref{table:1}. 

\paragraph{Experimental paradigm.} Participants in the first two studies were recruited via Amazon's Mechanical Turk \citep{buhrmester_amazons_2011}. Event segmentations for the "Pieman" story were obtained from $205$ participants that listened to the story twice and also segmented the story twice \citep{michelmann_moment-by-moment_2021}, i.e., data from two behavioral runs were available. The verbatim instruction for these participants was: "press the space bar every time when, in your judgment, one natural and meaningful unit ends and another begins". Event segmentation for "Monkey in the Middle" was obtained with the same instructions, however, batches of participants annotated $4$ sections of the story that were similar in length: All participants listened to the full story but only segmented their respective section; a total of $30$ participants segmented the first quarter, 35 participants segmented the second quarter, $31$ participants segmented the third quarter, and $42$ participants segmented the final quarter of the story \citep{kumar_bayesian_2022}. Event segmentation for "Tunnel Under the World" was obtained from $10$ participants that participated in person \citep{lositsky_neural_2016}. 

\paragraph{Human consensus estimation and sentence boundary alignment.} Aggregated event boundaries were derived for all data by computing smoothed averages across participants and detecting local peaks in the time course of the agreement. Specifically, individual participants' data were aggregated in response vectors at a millisecond resolution that were set to $1$ if a response was recorded within the surrounding second and to $0$ otherwise. The averaged response vector was then smoothed with a Gaussian kernel and thresholded. Within clusters of neighboring points, each cluster’s maximum was then taken as a local peak that indicated an event boundary \citep[see also: ][for detailed descriptions of boundary computation]{michelmann_moment-by-moment_2021, kumar_bayesian_2022}. Note that these local peaks represent the "consensus solution" of event boundaries. In the "Pieman" story $19$ event boundaries were detected on the first and $20$ event boundaries were detected on the second behavioral run. For the story "Monkey in the Middle" $43$ event boundaries were detected, and for "Tunnel Under the World" $30$ event boundaries were found (see table \ref{table:1}). For the purposes of this study, we attributed behavioral event boundaries (individual button presses or boundaries from aggregate responses) to the nearest sentence boundary. The distance to each sentence boundary was computed in milliseconds, where the time of each sentence boundary was based on the average between the offset of the last word of the previous sentence and the onset of the first word of the next sentence. On the individual subject level, participants marked on average $26.41$ sentence boundaries in the "Pieman" story as event boundaries on the first behavioral run ($SD = 17.71, min = 3, max = 86$) and found $24.48$ boundaries on the second run  ($SD = 16.16, min = 3, max = 72$). For "Monkey in the Middle" the average numbers of event boundaries in the four quarters were $44.03$ ($SD = 28.92, min =3, max = 107$), $29.97$ ($SD = 20.88, min = 3, max = 77$), $20.61$ ($SD = 14.86, min = 2, max =  68$), and $27.05$ ($SD = 22.14, min = 1, max = 88$), in "Tunnel Under the World", individual participants marked on average $72.5$ sentence boundaries as event boundaries ($SD = 33.99, min = 15, max = 133$).

Attributing button presses to sentence boundaries made it possible to directly compute a distance metric between human event segmentation and GPT-3-derived event segmentation that always coincided with sentence boundaries. Specifically, we computed the Hamming distance between the human and model-based boundary vectors. In order to additionally obtain a continuous measure of boundary detection in humans, the response vectors from individual participants were simply averaged (no smoothing was applied). Since the resulting ratio of participants that responded within a given second can be interpreted as a response probability, the log of that ratio was computed for comparison with the model's log-probability output. 
\begin{center}

\begin{table}[b!]
 \begin{tabular}{||c c c c||} 
 \hline
 Story &  N words & N events (consensus) & N events (individual button presses)\\ [0.5ex] 
 \hline\hline
 "Pieman (run 1)"  & 1137 & 19 & 26.41 (SD = 17.71, range =  3-86)\\ 
 "Pieman (run 2)"  &    & 20 & 24.48 (SD = 16.16, range =   3-72)\\ 
 \hline
 "Monkey"  & 6028 & 43 &  \\
 "Monkey (Q1)"  &   &   & 44.03 (SD = 28.92, range = 3-107) \\ 
 
 "Monkey (Q2)"  &   &   & 29.97 (SD = 20.88, range = 3-77)) \\ 
  
 "Monkey (Q3)"  &   &   & 20.61 (SD = 14.86, range = 2-68) \\ 
  
 "Monkey (Q4)"  &   &   & 27.05 (SD = 22.14, range = 1-88) \\ 
 \hline
 "Tunnel"  & 4418 & 30& 72.5 (SD = 33.99, range = 15-133) \\ [1ex] 
 \hline
\end{tabular}
\caption{Word count and number of behavioral events for each of the three stories. Note that annotations for "Pieman" were collected twice, and that "Monkey in the Middle" was collected across four sub-groups that annotated a quarter of the story each.}
\label{table:1}
\end{table}
\end{center}
\subsection{Statistical comparison of event annotations}
To compare the distances between event annotations at the sentence level, a vector was created that held zero for each sentence boundary that was not an event boundary, and one for each sentence boundary that was also an event boundary. Two event annotations were then compared by computing the Hamming distance between their respective vectors (i.e.,  the proportion of disagreeing components). Subsequently, the true distance was compared to the distribution of distances under $100,000$ random sentence permutations and a p-value was obtained as the proportion of distances that were smaller under random conditions. To compare the similarity between individual participants' segmentations and the consensus solution with the similarity between GPT-3-derived segmentations and the consensus solution, the Hamming distance between every participant's sentence-level segmentation and the sentence-level segmentation from the consensus solution was computed. Likewise, the Hamming distance between each GPT-derived solution and the consensus solution was computed. Finally, these Hamming distances were compared between GPT-3 and the human participants with an independent sample two-sided t-test. To compare the log-probability between the behavioral event annotations and the newline characters predicted by GPT-3, we first interpolated NaNs in the vector that held the model output. In the human data, the button-press probability was already defined at each time-point and could be log-transformed for comparison. However, there were some moments where no participant had pressed a button, resulting in time-points where the log-transformation of the probability is undefined. To handle these missing values, we therefore interpolated the log-probability of button press at those time points. Subsequently, the cross-correlation between the vectors was computed and the zero-lag correlation was tested statistically against non-correlation.



\section{Results}
\label{sec:results}

\subsection{Event segmentations from GPT-3 significantly align with consensus human annotations}
\begin{figure}[!ht]
 \centering
 \includegraphics[scale=0.56]{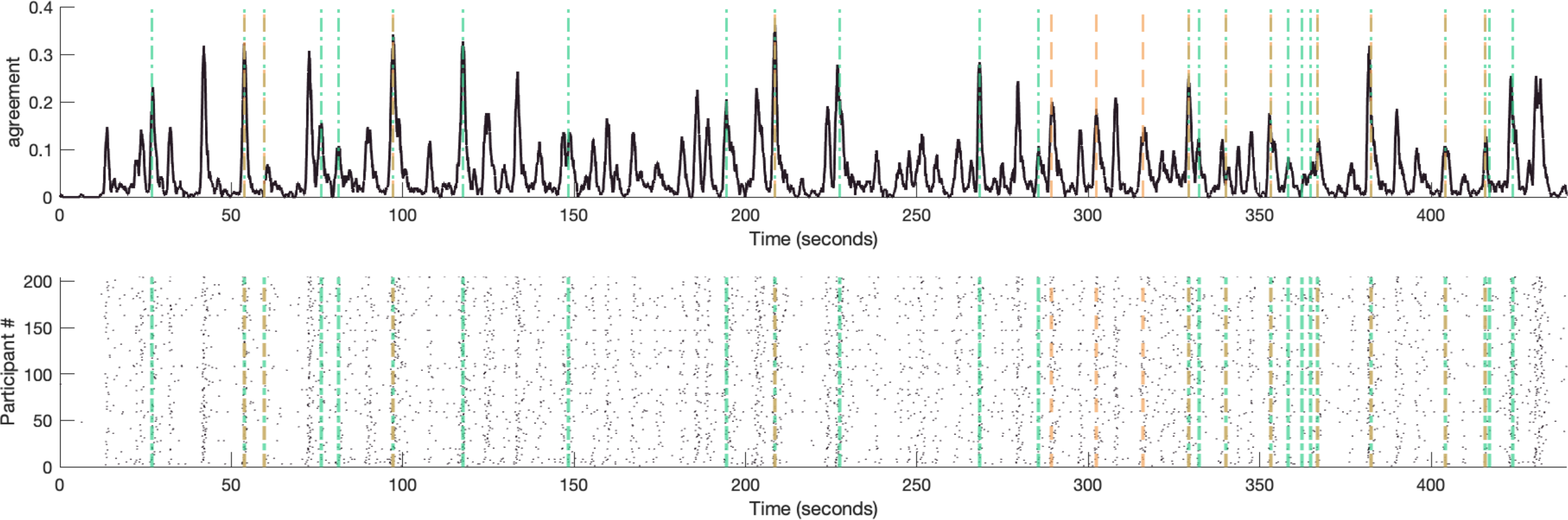}

  \caption{Example of human event boundary annotations and GPT-3-derived event boundaries. Vertical bars show event boundaries detected by GPT-3 with (orange) and without (green) the \textit{long} qualifier. The overlaid time-course of agreement (top) is from the second behavioral run of the "Pieman" story.  The overlaid raster plot (bottom) shows individual button presses of each participant.}
  \label{fig:ex}

\end{figure}

We segmented the text from the stories "Pieman", "Monkey in the Middle" and "Tunnel Under the World" into discrete events using GPT-3 (version "text-davinci-002"). "Pieman" was segmented into $23$ events by GPT-3 (compared to $19$ events in the averaged human annotations based on the first behavioral run and $20$ events in the annotations from the second behavioral run, see Figure \ref{fig:ex} for an example comparison of GPT-3-derived event boundaries and human button presses). For "Monkey in the Middle" we obtained $88$ events from the model output (compared to $43$ events in the averaged behavioral sample) and for "Tunnel Under the World", we obtained $139$ events (vs. $30$ behavioral events). Next, we computed the Hamming distance between the boundary-vectors that were derived from the GPT-3 model output and the boundary-vectors of annotations from the average human response. For "Pieman" we obtained a Hamming distance of $0.255$ when comparing to the human boundaries from the first behavioral run and a Hamming distance of $0.245$ when comparing to event boundaries from the second behavioral run. These distances were significantly smaller than what was obtained from comparisons under random permutation of sentence order ($p = 0.024$, and $p = 0.009$, respectively). For  "Monkey in the Middle" we obtained a Hamming distance of $0.25$ ($p = 0.044$) and "Tunnel Under the World" yielded a Hamming distance of $0.264$ ($p < 0.001$). 

The model found more event boundaries than the number of behavioral event boundaries in all stories. However, we also observed a wide range of individual human boundary annotations (see Section \ref{sec:methods}). We next set out to decrease the number of event boundaries in the model output by prompting the model to segment the story into \emph{long} events.

\paragraph{Segmenting \emph{long} events.} We observed that prompting GPT-3 to segment \emph{long} events results in segmentation that is even closer to the human consensus. Prompting the model to segment into \textit{long} events resulted in $14$ events in the "Pieman" story (v.s. $23$ without the \emph{long} qualifier), $59$ events in "Monkey in the Middle" (v.s. previously $88$ without the \emph{long} qualifier) and $76$ events in "Tunnel Under the World" (v.s. $139$). When comparing each model-derived event segmentation to human annotations, we obtained a Hamming distance of $0.223$ and $0.191$ with the "Pieman" story, when comparing to human annotations from the first and second behavioral run. These distances exceeded the distances that were observed under sentence level shuffle with a statistical trend compared to the first behavioral run ($p = 0.069$) and with statistical significance when compared to the second behavioral run ($p = 0.004$). For the story "Monkey in the Middle" we obtained a Hamming distance of $0.193$ ($p = 0.024$) and for "Tunnel Under the World" the Hamming distance between human annotations and the segmentation based on GPT-3 was $0.168$ ($p = 0.002$).

\begin{center}
\begin{table}[!ht]
 \begin{tabular}{||c c c c c||} 
 \hline
 Story &  N GPT events & distance / p-val & N GPT events (long) & distance / p-val\\ [0.5ex] 
 \hline\hline
 "Pieman (run 1)"  & 23 & 0.255 (p = 0.024)  & 14 & 0.223 (p = 0.069)  \\ 
"Pieman (run 2)"  &   &   0.245 (p = 0.009) &   &  0.191 (p = 0.004)\\  
 \hline
"Monkey"  & 88 & 0.25 (p = 0.044)& 59 & 0.193 (p = 0.024)\\
 \hline
 "Tunnel"  & 139 & 0.264 (p<0.001) & 76 & 0.168 (p = 0.002)\\  [1ex] 
 \hline
\end{tabular}
\caption{Number of events derived from GPT-3 output and the corresponding Hamming distance to the consensus human annotations.  The event boundaries segmented by GPT-3 significantly align with the consensus human annotations for almost all tested settings, as indicated by p-values $<0.05$.}
\label{table:2}
\end{table}
\end{center}

\paragraph{Reproducibility.} Because GPT-3 does not produce fully deterministic results (even with a \textit{temperature} setting of $0$), we next wanted to replicate our findings by repeating the model-based event segmentation five additional times for each story. With the "Pieman" story, we observed only one diverging replication where the model found $26$ events that had a Hamming distance of $0.287$ to the first ($p = 0.056$) and $0.255$ to the second  behavioral segmentation ($p = 0.006$) under the original event segmentation prompt. With "Monkey in the Middle" we observed two additional solutions that both had a Hamming distance of $0.25$ to the behavioral data with $86$ events ($p =  0.074$) and $85$ events ($p = 0.069$), "Tunnel Under the World" also yielded two additional solutions of $140$ and $139$ events that had Hamming distances of $0.266$  and $0.2622$ to the human annotations (all $ps < 0.001$). Note that the different solutions based on GPT-3 were very similar to each other with maximum Hamming distances between different solutions of $0.053$ for "Pieman", $0.042$ for "Monkey in the Middle", and $0.008$ for "Tunnel Under the world". When prompting the model to segment the stories into \textit{long} events, we also found diverging solutions in the five replications. For "Pieman" one additional solution yielded $14$ events and had a Hamming distance of $0.2446$ to human annotations from the first behavioral run ($p = 0.216$, n.s.) and $0.213$ to boundaries from the second run ($p = 0.022$). For "Monkey in the Middle" one additional solution with $56$ events had a Hamming distance of $0.186$ ($p = 0.016$) to the behavioral data. For "Tunnel Under the World" two additional solutions with $65$ and $68$ events were found that had Hamming distances of $0.145$ and $0.152$ to the behavioral data ($ps < 0.001$). Again, different solutions based on GPT-3 were very similar to each other with maximum Hamming distances between different solutions of $0.021$ for "Pieman", $0.007$ for "Monkey in the Middle", and $0.098$ for "Tunnel Under the World".

\subsection{GPT-3 event segmentation is statistically closer to the human consensus than individual humans}
\begin{figure}
 \centering
 \includegraphics[scale=0.34]{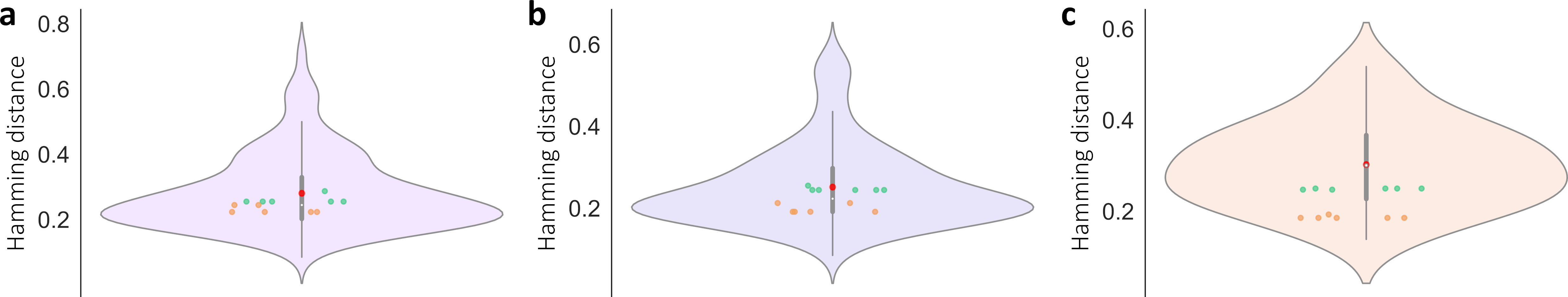}

  \caption{Distribution of distance to the consensus solution for human data and GPT-derived segmentation. Violin plots show the distribution of distances to the consensus for individual human button presses in the first (a) and second (b) behavioral run of the "Pieman" story, and for the story "Monkey in the Middle" (c). The thick gray bar represents the interquartile range, the red dot is the mean, the white dot is the median, and the thin gray line displays the extremes of the distribution excluding outliers. Colored dots represent the distance of the GPT-3-derived solution to the consensus, with (orange) and without (green) the \textit{long} qualifier.}
  \label{fig:vio}

\end{figure}

The significant similarities between GPT-3-derived event annotations and the event segmentation from human annotations suggest that GPT-3's segmentation reflects a shared human perception. Importantly, human-derived event boundaries are also a noisy approximation of an underlying shared event structure. Consequently, the question arises of how GPT-3 compares to typical human annotations that can be recorded in online experiments. In the "Pieman" story, for instance, GPT-3-derived event boundaries were consistently more similar to event boundaries from the second behavioral run compared to the first. In previous work, \cite{michelmann_moment-by-moment_2021} demonstrated that the second behavioral run is characterized by more agreement between participants, faster responses, and stronger correlations with neural measures of interest in electrocorticographic recordings from patients that listen to the story "Pieman". That is, the second behavioral run of "Pieman" may represent a better approximation of a perceived event structure. To test how GPT-3-derived event boundaries compare to individual human responses in measuring an underlying "consensus solution", we compared the average Hamming distance between GPT-3-derived solutions and the aggregate behavioral response (on the one hand) to the average Hamming distance between individual human participants' responses and the aggregate behavioral response (on the other hand). That is, we compared the similarity to the consensus solution between GPT-3 and individual human participants. We first tested this for the original event segmentation prompt (without the qualifier \textit{long}) for those stories that had enough annotations to derive a meaningful aggregate response ("Pieman": N = 205, and "Monkey in the Middle": N >= 30 per quarter, see Figure \ref{fig:vio}). For "Pieman" we found that individual participants' segmentations had an average Hamming distance of $0.281$ ($SD = 0.116$) to the consensus solution on the first behavioral run. The average Hamming distance of GPT-3-derived segmentations to the consensus solution was significantly lower ($0.261$; $SD = 0.012$, $p = 0.045$). On the second behavioral run, individual participants' segmentations had an average Hamming distance of $0.252$ ($SD = 0.099$) to the consensus solution; the GPT-3-derived segmentations still had a lower average Hamming distance ($0.246$, $SD = 0.004$), however, the difference was no longer significant ($p = 0.46$). Putting these results together: On the first run, GPT-3 does a better job, on average, of approximating the consensus than individual human participants;  on the second run, human participants increase their segmentation performance, so the GPT-3 segmentation is no longer significantly better. For "Monkey in the Middle", we treated concatenated response vectors from the four different segments (selected at random without replacement) as a single participant. These $30$ combined data sets had an average Hamming distance of $0.302$ ($SD = 0.094$) to the consensus solution (derived from the average across all participants). Again, the GPT-3-derived solutions were significantly more similar to the consensus solution with an average Hamming distance of $0.249$ ($SD = 0.001$, $p = 0.005$) supporting the idea that GPT-3-derived segmentations are closer to a consensus solution that large samples of human raters converge on (see Figure \ref{fig:vio}). 
We next repeated this analysis with the \textit{long} event segmentation prompt. For "Pieman" the average Hamming distance of \textit{long} GPT-3-derived segmentations to the consensus solution was $0.23$ ($SD = 0.01$) for the first behavioral run and $0.199$, ($SD = 0.01$) for the second behavioral run. Both of these values were significantly lower than the average for human annotations ($p < 0.001$ in both cases). For "Monkey in the Middle", the average Hamming distance of \textit{long} GPT-3-derived segmentations to the consensus solution was $0.187$ ($SD = 0.003$), which was significantly lower than the average for human annotations ($p < 0.001$). The lower Hamming distances under the \textit{long} event segmentation prompt suggest that these long events may be closer to shared human perception.

\subsection{Comparison of boundary probabilities between GPT-3 and human annotators}
\begin{figure}
 \centering
 \includegraphics[scale=0.46]{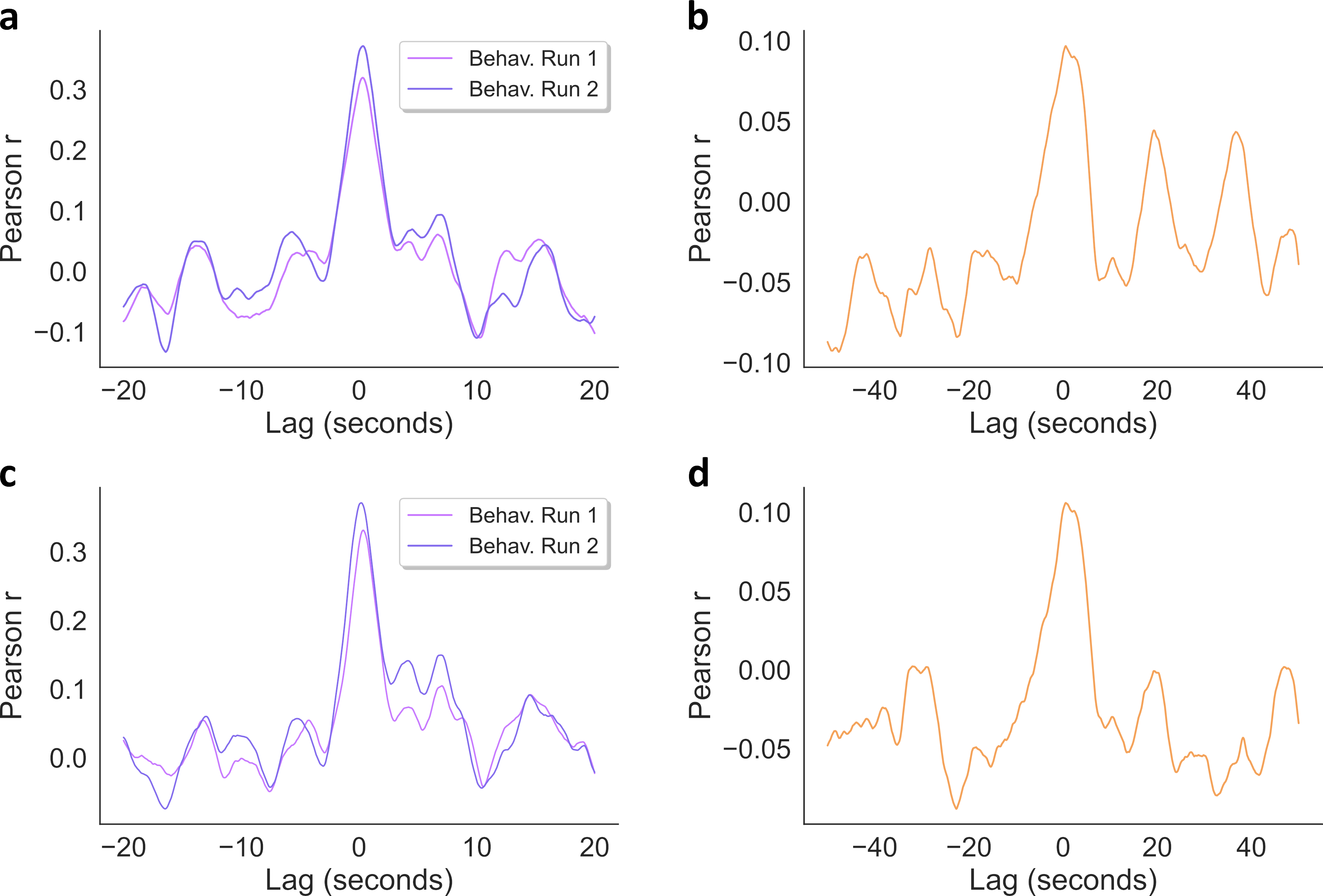}

  \caption{Cross-correlations between log-probability of button press in the behavioral samples and the log-probability of the newline character in the model output (a) "Pieman" original prompt (b) "Monkey in the Middle" original prompt (c) "Pieman" \textit{long} event prompt (d) "Monkey in the Middle" \textit{long} event prompt}
  \label{fig:cross}

\end{figure}

The event boundaries from human participants are those moments where a substantial amount of people pressed a button within a certain time window (i.e., peaks in agreement). That is, human event boundaries are discrete measures derived from a continuous variable (ratio of responses). Another way to treat this measure, however, is as the probability of button presses at every moment in the story. Similarly, predictions in GPT-3 can be treated as probabilistic. In order to compare the probability of event boundaries directly, we, therefore, log-transformed human event annotations to obtain the log-probability of a button press throughout the story. For GPT-3, we derived a continuous output variable from the log-probability of the newline character for every token in the story (averaged across all $6$ iterations of the model) and we interpolated those values onto the same time axis (see Section \ref{sec:methods} for detail). We then computed the cross-correlation between those time courses (note that we did not expect a specific lag between the time-courses, however, because human responses may entail a perception-to-action delay, we computed the cross-correlation to obtain an upper bound of the true correlation, see figure \ref{fig:cross}). With the original prompt, we obtained a zero-lag correlation of $r = 0.308$ ($p < 0.001$, maximum correlation of $r = 0.32$ at lag $303ms$) for the "Pieman" story's first behavioral run  and a zero-lag correlation of $r = 0.362$  ($p < 0.001$, maximum correlation of $r = 0.372$ at lag $330ms$) for the second behavioral run. For "Monkey in the Middle" we obtained a zero-lag correlation of $r = 0.092$ ($p < 0.001$, maximum correlation of $r = 0.097$ at lag $553ms$). Under the \textit{long} event prompt, we obtained a zero-lag correlation of $r = 0.311$ ($ p < 0.001$, maximum correlation of $r = 0.331$ at lag $362ms$) with the first behavioral run of the "Pieman" story and a zero-lag correlation of $r = 0.369$ ($p < 0.001$, maximum correlation of $r = 0.371$ at lag $169ms$) with the second behavioral run. For "Monkey in the Middle" we observed a zero-lag correlation of $r = 0.102$ ($p < 0.001$, maximum correlation of $r = 0.106$ at a lag of $552ms$). Taken together these findings are evidence that GPT-3 can represent the event structure of a naturalistic story in its predictions, rendering it a feasible tool for obtaining event annotations.

\clearpage

\section{Discussion}
\label{sec:discussion}


Here we demonstrate that GPT-3 can segment a continuous narrative into events, akin to the way human annotators perform this task. Notably, the present results may represent a lower bound of the similarity between GPT-3-derived annotations and human event perception, because human participants in our study segmented the audio form of the narrative, whereas GPT-3 only had access to the text itself. In the audio form of the narrative, the speaker may convey information about event structure via pauses and intonation that are absent from the text and they may obscure events by rapidly connecting consecutive sentences. Furthermore, GPT-3 performed event segmentation based on a full text-segment; therein, the model could use information from the current and next event to place an event boundary. Human annotators, on the other hand, had to make button-press decisions only based on past information (with the notable exemption of the second behavioral run of the "Pieman" story, where human participants could use their memory of the text to improve their segmentation). \\

A key benefit of using language models for event segmentation is that this approach is more scalable to very long naturalistic stimuli; as noted earlier, obtaining human-derived annotations for very long stimuli can be prohibitively labor- and resource-intensive. GPT-3 derived boundaries were also significantly closer to the ``consensus'' solution, on average, than boundaries identified by individual human annotators. These findings suggest that GPT-3 is a feasible method for cognitive scientists that are interested in event segmentation. In situations where individual human annotations are expected to be noisy and large behavioral samples are not available, a language model that has been trained on 45TB of text may be a useful alternative to obtain normative information about human text perception. Importantly, however, the variance between human annotators may contain meaningful information about individual human event perception; consequently, despite GPT-3's ability to capture human event perception, human behavioral data cannot be replaced as the gold standard for the investigation of human cognition.

\paragraph{Future work.} 
The demonstration that GPT-3 is able to segment a continuous narrative into events similarly to humans has implications for both cognitive science and machine learning. In cognitive science, researchers have already started viewing language models as useful \emph{model organisms} for human language processing \citep{tonevabridging} since they present one implementation of a language system that may be very different from that of the human brain, but may nonetheless offer insights into the nature of the linguistic tasks and the computational processes that are sufficient (or insufficient) to solve these tasks \citep{mccloskey1991networks,baroni2020linguistic}. Our findings suggest that a large language model can be viewed not only as a model organism for language processing but also more generally for event cognition. On the other hand, even though GPT-3-derived event segmentation significantly aligns with average human behavior, there is still a substantial gap between the behavior of the model and that of the most consistent humans. Biasing the language model to internally represent even more human-like event boundaries may result in improved performance on machine learning tasks. For instance, prior work has already demonstrated that "chunking" continuous input can help a modified transformer model to improve its memory performance \citep{lampinen_towards_2021}. Chunking narrative texts into events in a way that better mirrors human event perception may be a productive avenue to optimize cognitive tasks in machines. 

\section{Acknowledgements}
This work was funded by ONR MURI grant N00014-17-1-2961 awarded to KAN. MT was funded by the Princeton Neuroscience Institute CV Starr fellowship and NIMH T32 MH065214, and SM is funded by Deutsche Forschungsgemeinschaft (DFG), project 437219953. The authors wish to thank Shailee Jain, Vy Ai Vo, Javier Turek, and Alex Huth for helpful comments and discussions.


\bibliography{references}
\bibliographystyle{natbib}



\appendix
\section{Appendix}
\subsection{Anecdotal observations about the automated segmentation procedure}
A major challenge that we faced in finding a feasible approach to automated event segmentation was how to force the model to maintain an exact word-for-word copy of the story. Even with a \emph{temperature} setting of $0$, some randomness persists in the model output. With other approaches, the model would sometimes diverge from the text, change spelling, or even rephrase sentences. Crucially, the model could also start summarizing the events instead of providing a word-for-word copy. Key elements to solving this model behavior were: 
\begin{enumerate}
    \item Using the model version "text-davinci-002" over "text-davinci-003". The newer version tends to summarize events and ignores the instruction to provide a word-for-word copy.
    \item Using a succinct definition of "event" in the prompt.
    \item Adding a reminder about the instruction to copy word-for-word after the story segment in the prompt.
    \item Not using the full context window. Toward the end of the output, the model would diverge more often. Using a shorter context window helped with that.
    \item Keeping the \emph{max\_tokens} parameter low, but high enough to encompass a copy of the segment. 
    \item Having a clean transcription of the text. The model is most likely to diverge from the word-for-word copy when it corrects mistakes in the text. 
\end{enumerate}

Notably, a perfect word-for-word copy of the text is not a requirement for obtaining a meaningful event segmentation from the model, however, it may considerably increase the effort that is needed for automated parsing of the output. 

\subsection{Anecdotal observations on prompt variations}
The qualification of the term event can be adjusted further by asking for \emph{very long} events, or for \emph{extremely long} events, which results in coarser event segmentation in the model output. The definition of events as an "ongoing coherent situation" is in line with its description in the literature \citep[e.g.][]{zacks_event_2007}. Other definitions may work; however, making the prompt too verbose appears to result in a higher probability of text output that diverges from a word-for-word copy. Omitting the definition, on the other hand, results in a segmentation that appears to be based on the colloquial understanding of event as "occurrence". 

\subsection{Example of the Pieman story, segmented into events by GPT-3}
\itshape
I began my illustrious career in journalism in the Bronx where I toiled as a hard boiled reporter for The Ram, the student newspaper at Fordham University. \\

And one day I'm walking toward the campus center. And out comes the elusive Dean McGowan, architect of a policy to replace Fordham's traditionally working to middle class students with wealthier, more prestigious ones. So I whip out my notebook. And I go up to him and I say: Dean McGowan, is it true that Fordham University plans to raise tuition substantially above the inflation rate? And if so, wouldn't that be a betrayal of its mission? \\

And he stops. And looks at me. And he says: Listen up punk. And right then there's a blur in the corner of my eye which becomes this figure holding a cream pie which becomes the guy standing next to me mashing a cream pie into Dean McGowan's face. And and then runs away. And the Dean is covered with cream. \\

So I give him a moment. And then I say: Dean McGowan, would you care to comment on this latest attack? And the Dean says: Yes, I would care to comment. Fuck you. \\

So I race back to the newsroom with my scoop. And I, and I find the editor, Jim Dwyer, who's a senior. And he will go on to win a Pulitzer Prize. That is true. And he's a big guy. I pitch him my story. I tell him what I've seen. And he says: Dean McGowan? That guy's a dick! Write it up! \\

So I'm banging out my story. And I know it's good. And then I start to make it better by adding an element of embellishment. Reporters call this making shit up. And they recommend against crossing that line. But I had just seen the line crossed between a high powered dean and assault with a pastry. And I kind of liked it. So the first thing I did was, I gave the figure a name. I called him Pie Man, capital P, capital M. And I described him as a cape wearing, masked avenger, though in fact, he'd been capeless. And I said that as he fled the scene he clicked his heels in rakish glee. And I gave him a catchphrase in Latin. I said that he cried out: Ego sum non un bestia! Which means: I am not an animal. Which makes no sense. I needed something. I'm Catholic. Latin just comes to me. \\

So I finish my story and I hand it to Dwyer. And he reads. And he says: Pie Man, I love it! Page one. And that's how the first line got crossed. \\

A few days later I get a letter. It says: Dear Jim, good story. Nice details. If you want to see me again in action, be on the steps of Duane Library, Tuesday at three o'clock. Signed, Pie Man. \\

So I was there with a photographer from The Ram. And sure enough, out comes Sheila Beale, student body president. And now Sheila was different from me and all the other Fordham students who wore flannel shirts and worked part time jobs. Sheila was well bred. Sheila had school spirit. Sheila was the kind of student that Dean McGowan wanted more of. Although rumor had it that he got plenty of her in his office on his desk. Oh, but that's just a rumor. Please do not spread it outside this room. I myself would never say that. But the fact was Sheila had collaborated with the Dean to ban outdoor drinking on campus. The infamous no more beer at barbecues rule. That's right, boo that rule! Sheila thought drinking in public was in poor taste. \\

I think you know what happened next. Pie Man emerged from behind the late night library drop box, made his delivery, and fled away crying: Ego sum non un bestia. Or at least, that's what it said in my story in the newspaper the next day, which ran with a photo of him leaving the scene cape flowing behind him doing this. And that's what made him a sensation on campus. People started dressing like him and and quoting him in class. The Ram ran five major stories about Pie Man, all of them by me. \\

And toward the end of this run I was out at a bar one night. \\

And I saw, I came in I saw, in the corner, Angela from my Brit Lit class drinking with some friends. \\

And now, Angela and I had been flirting for two months. Or, I had been flirting with her. And with such nuance that there was a question about whether she knew I existed. \\

So I I saw her there and made a mental note to do nothing about it. \\

And then I went to the bar to buy a round. \\

And I felt a tap on my shoulder. \\

And it was her. \\

And she said: Jim, we were just talking about how you always seem to know when and where Pie Man will strike. And we were wondering. Are you Pie Man? \\

And I knew, by the way she said it, I knew that if I said yes, she would have sex with me. And wasn't I really Pie Man for having brought him into being? Didn't she only know about him through me? \\

But she had asked me a straightforward question that came with a straightforward answer. In fact, I wasn't Pie Man. As far as I knew, I had never seen the guy out of costume. \\

So I looked at her. \\

And I said: Yes, Angela, I am Pie Man. \\

And she said: Oh, good. Now buy me a beer and tell me all about it.

\end{document}